\documentclass[journal]{IEEEtran}

\usepackage{graphicx}
\usepackage{epstopdf} 
\graphicspath{{./figures/}}

\usepackage{hyperref}

\usepackage{algorithm}
\usepackage{algpseudocode}

\usepackage{amsmath}
\usepackage{amsthm}
\usepackage{amsfonts}
\usepackage{amssymb}

\usepackage{multirow}

\usepackage[utf8]{inputenc}

\usepackage[left=0.62in,right=0.62in,top=0.75in, bottom=1in]{geometry}

\usepackage{color}

\begin{document}
\pagestyle{empty} 

\title{The Future of Aerial Communications: A Survey of IRS-Enhanced UAV Communication Technologies}

\author{\IEEEauthorblockN{Zina Chkirbene$^1$, Ala Gouissem$^2$, Ridha Hamila$^1$, Devrim Unal$^3$}\\
\IEEEauthorblockA{$^1$Electrical Engineering, Qatar University, Qatar}\\
\IEEEauthorblockA{$^2$College of Computing and Information Technology, University of Doha for Science and Technology, Qatar}\\
\IEEEauthorblockA{$^3$KINDI Center for Computing Research, College of Engineering, Qatar University, Qatar}
}

\maketitle

\begin{abstract}
The advent of Intelligent Reflecting Surfaces (IRS) and Unmanned Aerial Vehicles (UAVs) is setting a new benchmark in the field of wireless communications. IRS, with their groundbreaking ability to manipulate electromagnetic waves, have opened avenues for substantial enhancements in signal quality, network efficiency, and spectral usage. These surfaces dynamically reconfigure the propagation environment, leading to optimized signal paths and reduced interference. Concurrently, UAVs have emerged as dynamic, versatile elements within communication networks, offering high mobility and the ability to access and enhance coverage in areas where traditional, fixed infrastructure falls short. This paper presents a comprehensive survey on the synergistic integration of IRS and UAVs in wireless networks, highlighting how this innovative combination substantially boosts network performance, particularly in terms of security, energy efficiency, and reliability. The versatility of UAVs, combined with the signal-manipulating prowess of IRS, creates a potent solution for overcoming the limitations of conventional communication setups, especially in challenging and underserved environments. Furthermore, the survey delves into the cutting-edge realm of Machine Learning (ML), exploring its role in the strategic deployment and operational optimization of UAVs equipped with IRS. The paper also underscores the latest research and practical advancements in this field, providing insights into real-world applications and experimental setups. It concludes by discussing the future prospects and potential directions for this emerging technology, positioning the IRS-UAV integration as a transformative force in the landscape of next-generation wireless communications.

\end{abstract}

\section{Introduction}

Unmanned Aerial Vehicles (UAVs) and Intelligent Reflecting Surfaces (IRS) are transforming wireless communication systems, signifying a major technological advancement. UAVs, initially deployed for military applications, now serve in diverse commercial and research roles due to their autonomous operation and remote-control capabilities. They can function as mobile base stations or communication relays, which enhances network reach, throughput, and reliability \cite{azari2020uav}. Their versatility extends to advanced data collection, surveillance, and load balancing \cite{ge2023intelligent}. Meanwhile, IRS technology introduces a new paradigm with programmable environments composed of passive elements capable of modulating electromagnetic waves in real-time \cite{zeng2023irs}. This innovation optimizes radio signal propagation by reducing interference, improving signal quality, and overcoming multi-path fading \cite{samara2019performance}.

The convergence of UAVs and IRS technology has paved the way for more adaptable and efficient wireless networks. This combination of UAVs' mobility and IRSs' signal optimization creates a dynamic and flexible wireless landscape. The integration enables networks to overcome signal blockage, coverage limitations, and connectivity issues, especially in areas underserved by conventional infrastructure \cite{zhang2023joint}. It also provides fine-tuned solutions to problems like interference management and signal fading \cite{kang2023active}. However, merging UAVs with IRS introduces security challenges due to the airborne nature of UAVs and the passive techniques used by IRS \cite{an2023robust}.

Security in this combined technological environment is crucial. Ensuring the secrecy of UAV-to-UAV communication involves addressing large-scale fading channels and detecting UAV eavesdroppers \cite{ye2018secure} \cite{wang2020detection}. Proposed techniques to augment UAV security include jamming \cite{chkirbene2023secure}, trajectory alterations \cite{li2021robust}, and artificial noise \cite{li2021joint}. A notable method involves using IRS to improve secure communication \cite{yang2020secrecy}. By adjusting phase shifts, IRS can boost the signal-to-noise ratio for legitimate users while diminishing it for eavesdroppers, creating a potent strategy to enhance communication security without increasing power allocation \cite{wang2021secrecy}. As research in this field progresses, the challenge remains to optimize the phase shifts of IRS and the trajectories of UAVs to maintain robust and secure wireless networks.

Deploying these technologies, however, presents its own set of challenges, particularly in the realms of security, energy efficiency, and the requirement for intelligent management systems. The security of UAV-assisted communication networks is paramount, particularly in sensitive applications. Measures must be taken to ensure data integrity, prevent unauthorized access, and protect against cyber threats to maintain the reliability of these systems. Additionally, the energy efficiency of both UAVs and IRS is of critical importance, as the sustainability of these technologies relies on their ability to operate with minimal energy consumption while still maintaining high performance. Addressing these challenges, machine learning (ML) plays an indispensable role. ML algorithms can significantly optimize the performance of IRS-assisted UAV systems by intelligently managing resources, enhancing signal processing, and ensuring secure communications. 

Various surveys have explored IRS-assisted UAV communication systems, each with their specific focus and methodologies. Some works delve deeply into IRS technologies, while \cite{oubbati2020softwarization} focus more on the UAV aspect. Methodologically, surveys like \cite{pogaku2022uav} lean towards optimization algorithms. In terms of security and privacy, may emphasize technical performance or deployment strategies more than security aspects, and in the realm of energy efficiency, studies like \cite{pandey2022security} do not prioritize it as a primary focus.

This paper presents a distinct approach in the landscape of IRS-assisted UAV communication systems. It offers a comprehensive overview, emphasizing secure, energy-efficient IRS-assisted UAV systems with a special focus on machine learning. This survey not only highlights the technological synergy between UAVs and IRS but also integrates machine learning as a core methodology, setting it apart from the aforementioned studies. The survey addresses specific challenges and future research directions related to security, energy, and machine learning implementation, offering insights into areas less explored in other surveys. The integration of machine learning, particularly for optimizing system performance and security, underscores the unique contribution of this survey to the existing body of literature.


The paper is structured as follows:. Section II covers energy efficiency, detailing models for optimizing energy use while ensuring high performance.  Section III discusses security in IRS-assisted UAV communications, analyzing challenges and solutions. Section IV explores the role of Machine Learning (ML) in enhancing the IRS-UAV framework.  Section V summarizes key findings and future prospects in IRS-assisted UAV communications. Section VI, concludes the paper.

\section{Energy Efficiency in IRS-Assisted UAV Communication Systems}

Energy efficiency is a critical factor in the implementation of all systems, particularly in wireless communications where it greatly influences system performance and device size. As such, energy efficiency is considered one of the most fundamental performance criteria, alongside spectral efficiency. This section introduces research trends in IRS-assisted UAV communications systems from the perspective of energy efficiency.

In the realm of IRS-assisted UAV communications, significant research has been conducted to optimize energy efficiency. For instance, \cite{ge2020joint} investigated how to maximize the received power of ground users using a single UAV and multiple IRSs, revealing that the energy efficiency of such systems is highly dependent on the number of IRS elements. 
Another critical research, \cite{liu2020machine}, aimed to minimize UAV power consumption using a Decaying Deep Q-Network (D-DQN), which is an adaptation of the Deep Q-Network (DQN) algorithm with a decaying learning rate. This study was pivotal in comparing the performance of Orthogonal Multiple Access (OMA) and Non-Orthogonal Multiple Access (NOMA) systems in terms of energy efficiency.

The pursuit of enhanced energy efficiency has also led to novel approaches in multicell uplink systems using aerial base stations, particularly UAV-mounted IRS systems. 
In \cite{mei2021joint}, the optimization of energy efficiency in IRS-assisted UAV communication systems is discussed.  However, these UAVs often face challenges like obstructed UAV-to-ground links, leading to increased task latency. To address this, configuring the IRS to improve the propagation channels between the UAV and ground terminals becomes essential, thereby enhancing MEC performance and wireless communications. 
Similarly, \cite{cai2020resource} explored resource allocation strategies involving UAV trajectory and velocity to minimize the system's average total power consumption.
Moreover, \cite{al2022ris} aimed to enhance UAV connectivity and energy efficiency by maximizing the number of connectable devices during activation periods, showcasing the effectiveness of large-sized IRSs. \cite{michailidis2021energy} proposed an optimization strategy to minimize the weighted total energy consumption of vehicles, incorporating transmit power constraints and aerial roadside units within an IRS network.

\begin{table}[H]
\centering
\caption{Summary of previous works for energy efficiency. }
\label{security_improvement}
\begin{tabular}{|p{0.5cm}|p{1.7cm}|p{5.3 cm}|}
\hline
\textbf{Ref} & \textbf{Objective} & \textbf{Design Variables} \\ \hline
\cite{ge2020joint} & Maximize the received power at the ground users & \textbf{UAV Active Beamforming}: Dynamic adjustment of UAV signal transmission. \textbf{IRS Passive Beamforming}: Signal reflection optimization by IRS. \textbf{UAV Trajectory}: Flight path optimization for signal transmission. \\ \hline
\cite{liu2020machine} & Minimize the energy consumption of UAV & \textbf{Movement of UAV}: Energy-efficient UAV flight path adjustments. \textbf{IRS Phase Shift}: IRS signal phase manipulation. \textbf{Power Allocation Policy}: Strategic power distribution for energy conservation. \\ \hline\cite{mei2021joint} & Maximize the energy efficiency & \textbf{UAV Trajectory}: Path planning for efficiency. \textbf{Task Offloading}: Distribution of tasks between UAV and ground stations. \textbf{Cache with Phase Shift Design of IRS}: Data caching optimization alongside IRS adjustments. \\ \hline
\cite{long2020joint} & Maximize the secrecy energy efficiency & \textbf{UAV Trajectory}: Path adjustment for secure communication. \textbf{IRS Phase Shift}: Signal security enhancement through IRS. \textbf{User Allocation}: Optimal user positioning for security. \textbf{Transmit Power Control}: Power management for secure transmission. \\ \hline
\cite{cai2020resource} & Minimize the average total power consumption of the system & \textbf{Resource Allocation Strategy}: Optimal distribution of system resources. \textbf{UAV Trajectory}: Efficient UAV path planning. \textbf{UAV Velocity}: Speed optimization for energy saving. \textbf{IRS Phase Shift}: Adjustment of IRS signal reflection for power reduction. \\ \hline
\cite{al2022ris} & Maximize the number of available devices & \textbf{UAV Trajectory}: Strategic UAV flight path for device connectivity. \textbf{IRS Phase Shift}: IRS adjustments for device coverage. \textbf{User Allocation and Transmit Power}: User positioning and power management for maximum device availability. \\ \hline
\cite{khalili2021resource} & Minimize the total transmit power & \textbf{UAV Trajectory and Velocity}: Path and speed optimization for power saving. \textbf{IRS Phase Shift}: IRS signal manipulation for reduced transmission power. \textbf{Subcarrier Allocation}: Efficient distribution of frequency subcarriers. \\ \hline
\cite{yang2020performance} & Improve the coverage and reliability of UAV communications & \textbf{Outage Probability}: Reduction in communication failures. \textbf{Average Bit Error Rate (BER)}: Improvement in signal accuracy. \textbf{Average Capacity}: Enhancement of data transmission capability. \\ \hline
\end{tabular}
\end{table}
In \cite{cang2020optimal}, the authors focused on reducing UAV transmit power through strategic placement, IRS phase shift adjustments, user association, and IRS association. This approach included a novel framework using visible-light-communication and extensive IRS deployment, showing a significant reduction in energy consumption. Finally, \cite{khalili2021resource} presented a novel approach for heterogeneous networks (HetNets) supporting dual connectivity through the use of multiple UAVs as manual relays equipped with IRSs. And \cite{yang2020performance} investigated methods to enhance coverage and reliability in UAV communication systems with IRS assistance, developing an approach to model the statistical distribution of the ground-to-air link signal-to-noise ratio (SNR).

The summarized research works in the table \ref{security_improvement} collectively focus on enhancing the performance and efficiency of UAV-based communication systems, often utilizing IRS technology. These studies primarily aim to maximize signal strength, energy efficiency, and security while minimizing power consumption. Key strategies include optimizing UAV trajectories, employing active and passive beamforming techniques, and manipulating IRS phase shifts. Several works also emphasize the importance of resource allocation, user positioning, and task management to improve overall system effectiveness. These contributions highlight the growing significance of UAVs and IRS in advancing modern communication technologies, offering insights into optimizing these systems for better coverage, reliability, and energy efficiency.
\section{Enhancing Security in IRS-Assisted UAV Communications}
In the dynamic and ever-evolving domain of wireless communications, the integration of IRS with UAVs has emerged as a pivotal innovation, opening new frontiers in enhancing network capabilities. A critical dimension of this advancement is the bolstering of security in IRS-assisted UAV communication systems. As wireless networks grow in complexity and ubiquity \cite{chkirbene2020smart}, the implementation of robust security measures becomes increasingly crucial to safeguard against a diverse range of cyber threats and vulnerabilities.
This section delves into the sophisticated methodologies and strategic approaches employed to strengthen the security of IRS-assisted UAV communications. In \cite{sun2021intelligent,fang2021secure}, the authors have investigated the strategic positioning of UAV Base Stations (BSs) and IRSs. These efforts focus on navigating the challenges arising from the presence of both legitimate and illegitimate interceptors, with an aim to fortify the confidentiality of transmissions. This involves optimizing the configuration of the beamforming weight matrix to elevate the secrecy rate of communications.
Building on these insights, research such as \cite{fang2020joint} has introduced innovative power control strategies for UAVs. These strategies are designed to further increase the average secrecy rate, thereby enhancing the security of data transmissions. Additionally, \cite{sun2021intelligent} explores the probability of establishing Line-of-Sight (LoS) links within UAV communication systems, paying special attention to the impact of IRS deployment on these probabilities.
In tackling the challenge of secure data transmission within IRS-supported UAV millimeter-wave (mmWave) communication networks, \cite{guo2021learning} adopts a Deep Deterministic Policy Gradient (DDPG) methodology. This approach aims to maximize the cumulative secrecy rate for all authorized users, introducing an innovative Dual Deep Reinforcement Learning (DRL) algorithm to efficiently achieve this objective. 
Moreover, in \cite{li2021robust} a new model that takes a holistic approach by concurrently optimizing UAV flight paths, IRS-assisted passive beamforming, and controlled transmission power for authorized communications. This comprehensive strategy focuses on enhancing the average worst-case secrecy rate, thereby contributing significantly to the security of IRS-assisted UAV systems. 
\begin{table}[h!]
\centering
\caption{Summary of previous works relevant to the improvement of security.}
\label{sp}
\begin{tabular}{|p{0.5cm}|p{2.5cm}|p{4.5cm}|}
\hline
\textbf{Ref} & \textbf{Objective} & \textbf{Design Variables} \\ \hline
\cite{sun2021intelligent} & Maximize the secrecy rate & \textbf{UAV-BS and IRS Positions}: Optimal placement of UAV-based stations (BS) and IRS for enhanced security. \textbf{UAV-BS and IRS Beamforming}: Using beamforming techniques at both UAV-BS and IRS to strengthen secure communications. \\ \hline
\cite{fang2021secure} & Maximize the secrecy rate & \textbf{UAV Location}: Strategic positioning of the UAV to optimize signal secrecy. \textbf{IRS Phase Shift}: Adjusting the phase shift of the IRS to enhance communication security. \\ \hline
\cite{fang2020joint} & Maximize the average secrecy rate & \textbf{IRS Phase Shift}: Fine-tuning IRS phase shifts for secure transmissions. \textbf{UAV Trajectory}: Planning the UAV's flight path to maximize secrecy rate. \textbf{Transmit Power of UAV}: Adjusting UAV's transmission power for optimal secrecy. \\ \hline
\cite{guo2021learning}& Maximize the sum secrecy rate of all legitimate users & \textbf{UAV Trajectory}: Determining the UAV’s flight path to enhance user secrecy rates. \textbf{Coefficients of the IRS Elements}: Optimizing IRS elements for maximum security. \textbf{UAV Active Beamforming}: Dynamic beamforming to improve secure communication. \\ \hline
\cite{li2021robust} & Maximize the worst-case secrecy rate & \textbf{UAV Trajectory}: Flight path optimization for worst-case security scenarios. \textbf{IRS Passive Beamforming}: IRS adjustments to ensure robust secrecy rates. \textbf{Transmit Power of the Legitimate Transmitters}: Power control of legitimate transmitters to enhance secrecy against eavesdroppers. \\ \hline
\end{tabular}
\end{table}

The studies summarized in Table \ref{sp} collectively emphasize enhancing the security of UAV-based communication systems using IRS technology. Key focuses include optimizing UAV and IRS positioning, fine-tuning beamforming techniques, and adjusting IRS phase shifts to maximize secrecy rates. These works demonstrate a diverse range of strategies, from trajectory planning and power control to securing communications against eavesdropping, highlighting the significance of UAV and IRS integration in strengthening the security aspect of modern wireless communication networks.

\section{Leveraging AI in IRS-Assisted UAV Communications}

The convergence of IRS and UAVs represents a landmark advancement in various applications, particularly enhancing UAV communications. This amalgamation, further boosted by the integration of Artificial Intelligence (AI) and Machine Learning (ML), catalyzes significant improvements in communication quality. It marks a pivotal step in elevating aspects such as Quality of Service (QoS), reliability, security, and overall network performance, signaling a new epoch in intelligent communication technologies.

IRS-assisted UAVs emerge as a vanguard of intelligent decision-making, adept at sophisticated data analysis, predictive analytics, and navigating complex scenarios to achieve optimal outcomes. Machine Learning plays an integral role in this advancement \cite{chkirbene2020weighted}. Its various forms, including supervised, unsupervised, and reinforcement learning, are crucial in augmenting channel estimation, system optimization, UAV tracking accuracy, spectral efficiency, and effectively managing operational trade-offs. These ML algorithms are central to meticulously fine-tuning UAV positions and IRS phase shifts, thereby enhancing communication outcomes. They underpin accurate tracking and maintaining reliable channel conditions, which are vital for high-performance standards. Advanced methodologies like Convolutional Neural Networks (CNNs) and Deep Neural Networks (DNNs) are employed to reduce system complexity and computational demands significantly \cite{guizani2020combating}.

Deep Learning (DL) is increasingly becoming a focal point in IRS-assisted UAV communication research. Present studies are exploring the application of deep learning algorithms to improve the efficiency, reliability, and adaptability of IRS-assisted UAV systems, marking a significant stride in next-generation communication technologies. For instance, in \cite{9743298}, a study on UAV-aided wireless communications applies a three-dimensional dynamic channel model in IRS-assisted UAV communication for the first time. It introduces a novel deep learning-based channel tracking algorithm that integrates a pre-estimation module with denoising deep neural networks and a channel tracking module using stacked bi-directional long short-term memory (Stacked Bi-LSTM) networks.

The interest in DL-enabled solutions for metasurface reconfiguration \cite{gouissem2022federated} and deep reinforcement learning (DRL) algorithms in IRS-empowered wireless networks and UAV-supported networks is growing. Studies including \cite{ nguyen2020real} are pioneering new solutions to optimize IRS phase shifts and enhance UAV performance. However, these studies often assume idealized conditions such as static environments or perfect Channel State Information (CSI), which may not reflect real-world complexities.

In response, more sophisticated techniques are being developed. The deep deterministic policy gradient (DDPG) algorithm, focuses on IRS-UAV NOMA downlink systems to optimize various system parameters. Concurrently, innovative DRL methodologies and adaptive IRS-assisted transmission strategies, as explored in \cite{mei20223d, cao2021reconfigurable}, are proposed to enhance communication quality and system throughput while optimizing power consumption.
Table \ref{MLresearch_comparison} presents a summary of these research works in IRS-assisted UAV communications.

\begin{table}[h!]
\centering
\caption{Comprehensive Comparison of Research Works in IRS-Assisted UAV Communications}
\label{MLresearch_comparison}
\begin{tabular}{|p{0.3cm}|p{1.3cm}|p{3cm}|p{2.7cm}|}
\hline
\textbf{Ref} & \textbf{Focus} & \textbf{Detailed Summary} & \textbf{Distinctive Aspect} \\
\hline
\cite{feng2020deep} & DL solutions in IRS-empowered wireless networks & Investigated deep learning for reconfiguring metasurfaces and developed deep reinforcement learning algorithms for network optimization. & Emphasized the use of deep learning for metasurface configuration, a novel approach at the intersection of AI and physical network optimization. \\
\hline
\cite{nguyen2020real} & Optimization in UAV-supported networks & Focused on optimizing IRS phase shifts to enhance UAV network performance, addressing signal quality and network reliability. & Centered on practical optimization strategies for IRS in UAV networks, specifically targeting the enhancement of signal reliability and strength. \\
\hline
\cite{huang2020reconfigurable} & IRS phase shift optimization & Examined the potential of IRS phase shifts in improving UAV communication, with a focus on signal quality and strength. & Similar to \cite{nguyen2020real} in optimizing IRS phase shifts but with a more focused examination of IRS's potential in signal enhancement. \\
\hline
\cite{samir2021optimizing} & Enhancing UAV performance & Studied the optimization of IRS phase shifts for UAV systems, aiming at improved communication and operational efficiency. & Although it also focuses on IRS phase shifts like \cite{nguyen2020real} and \cite{huang2020reconfigurable}, it leans more towards operational efficiency in UAV systems. \\
\hline
\cite{yu2022deep} & IRS-UAV NOMA downlink systems & Implemented the DDPG algorithm for optimizing various system parameters in IRS-UAV NOMA downlink systems. & Unique for its application of the DDPG algorithm in the specific context of IRS-UAV NOMA systems, emphasizing machine learning in system parameter optimization. \\
\hline
\cite{mei20223d} & Innovative DRL methodologies & Proposed new DRL approaches for IRS-assisted UAV communications, introducing cutting-edge methodologies. & Stands out for its innovation in DRL methodologies, specifically tailored for IRS-assisted UAV communication challenges. \\
\hline
\cite{cao2021reconfigurable} & Adaptive IRS-assisted transmission strategies & Developed strategies to improve communication quality and system throughput in IRS-assisted transmissions. & Focuses on adaptive strategies, differentiating itself by emphasizing adaptability and throughput in IRS-assisted communications. \\
\hline
\end{tabular}
\end{table}

\section{Performance evaluation}

In this simulation, we explore three distinct scenarios to assess the impact of Intelligent Reflecting Surfaces (IRS) and Unmanned Aerial Vehicles (UAVs) on wireless communication systems. The first scenario serves as a baseline, where communication occurs without the assistance of IRS, reflecting typical wireless network conditions. The second scenario introduces IRS into the system, aiming to enhance the communication link by leveraging the IRS's ability to smartly reflect and manipulate electromagnetic waves, thereby improving signal strength and coverage. The third and most complex scenario integrates a UAV equipped with IRS, positioning it optimally in the communication path. This UAV-assisted IRS setup is designed to dynamically alter the propagation environment, potentially overcoming obstacles and further enhancing signal quality and network performance over greater distances. 

\subsection{Simulation results}

\begin{figure}[h]
\centering
\includegraphics[width=3.6in,height=1.8in]{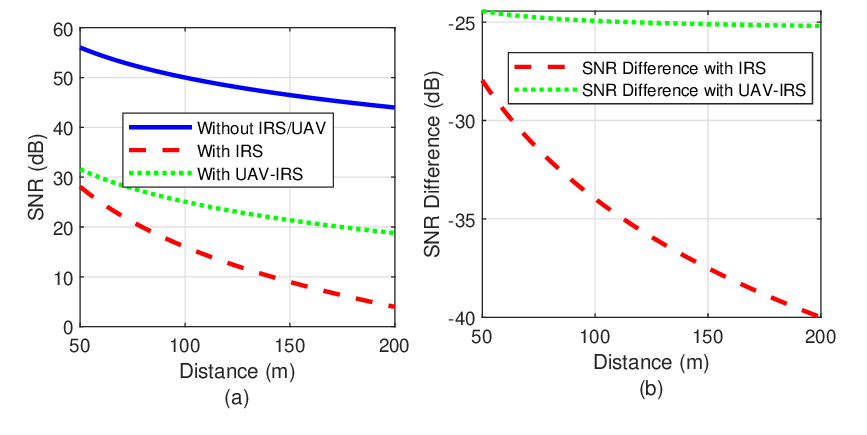} 
\caption{SNR performance of the system without IRS, with IRS and with UAV/IRS }
\label{fig1}
\end{figure}

Figure \ref{fig1} (a) shows the SNR performance across different distances for each system. Notably, the UAV-IRS system consistently outperforms the others, especially at shorter distances, benefiting from the UAV's enhanced gain. The IRS system also shows an improvement over the baseline system but is not as pronounced as the UAV-IRS system. Figure \ref{fig1} (b) presents the SNR enhancements offered by the IRS and UAV-IRS systems relative to the baseline system. It emphasizes the value added by the IRS and UAV-IRS technologies in terms of improved signal quality, offering a comparative perspective on their performance gains

\begin{figure}[h]
\centering
\includegraphics[width=3.6in,height=1.8in]{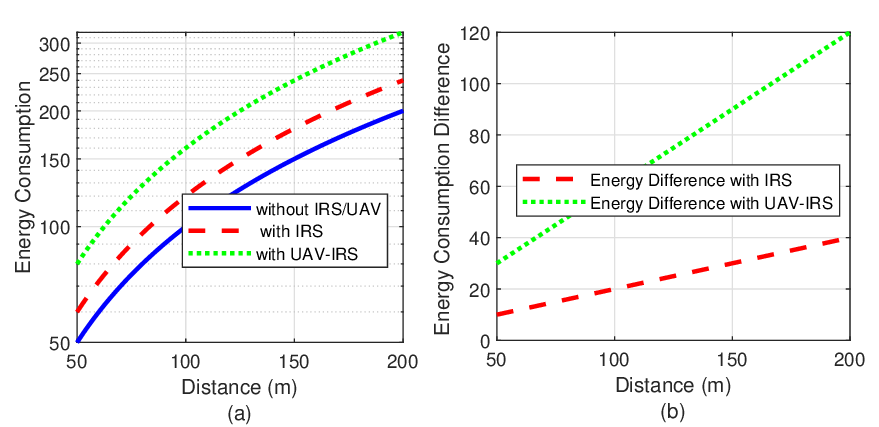} 
\caption{The energy consumption performance of the system without IRS, with IRS and with UAV/IRS }
\label{fig2}
\end{figure}

Figure \ref{fig2} (a) reveals the progressive energy consumption of each system as the distance increases. This figure indicates that the UAV-IRS system consumes the most energy, followed by the IRS system, and finally the baseline system. This outcome is expected due to the additional technology and functionality incorporated in the IRS and UAV-IRS systems. Figure \ref{fig2} (b) highlights the incremental energy consumption of the IRS and UAV-IRS systems compared to the baseline. It visually depicts the extra energy required for the advanced functionalities of IRS and UAV-IRS, providing a clear understanding of the energy trade-offs involved.
\section{Discussion}
The integration of IRS with UAVs in wireless communication systems, as explored in this survey, opens up a new paradigm in achieving advanced network performance, security, and energy efficiency. In Section II, the focus on energy efficiency revealed the critical importance of developing sustainable models for IRS-assisted UAV systems. Energy efficiency is not just a matter of environmental concern but also a practical necessity for the long-term operational viability of these systems. The models discussed underscore the need for innovative approaches to reduce power consumption while maximizing system performance, which is particularly challenging in scenarios involving extensive data transmission and processing. In Section III, we discussed the intricate security challenges that arise in IRS-assisted UAV communications. The unique vulnerabilities of these systems necessitate robust security protocols and innovative countermeasures to protect against evolving cyber threats. The development and implementation of these security measures are crucial for ensuring the reliability and integrity of data transmission in increasingly complex network environments.
The role of Machine Learning (ML), as examined in Section IV, is one of the most transformative aspect of IRS-assisted UAV communications. ML algorithms, including deep learning and reinforcement learning, show promise in optimizing system configurations, enhancing signal processing, and automating security responses. These advancements suggest a future where ML not only improves operational efficiency but also becomes integral to proactive security management and real-time decision-making in complex network scenarios. 

\section{Conclusion}
This survey has presented a thorough exploration of IRS-assisted UAV communication systems, emphasizing the significant innovations and challenges in security, energy efficiency, and the integration of machine learning (ML). The integration of UAVs with IRS technology marks a transformative development in wireless communications, offering enhanced flexibility and novel signal optimization capabilities. However, this convergence also introduces complex challenges that necessitate ongoing research and development.
IRS-assisted UAV communication systems stand at the forefront of wireless communication innovation. The path forward involves not only addressing the current challenges but also unlocking the full potential of these technologies. This realm offers expansive opportunities for future exploration and advancement, with the potential to significantly enhance telecommunications, surveillance, and data collection capabilities in modern communication networks.

\section*{Acknowledgment}

This work was supported by Qatar University Internal Grant  IRCC‐2023‐237.
The statements made herein are solely the responsibility of the author[s].

\bibliographystyle{IEEEbib}
\bibliography{Mybiblio}

\end{document}